\documentclass[letterpaper, 10 pt, conference]{ieeeconf}
\IEEEoverridecommandlockouts                   
\usepackage{amssymb} 
\usepackage{amsmath} 
\usepackage{blkarray} 
\usepackage{wrapfig} 
\usepackage{array}
\usepackage{graphics} 
\usepackage{graphicx,subfigure}
\usepackage{epsfig}
\usepackage[ruled,linesnumbered,vlined]{algorithm2e}
\usepackage{multirow}
\usepackage{caption}
\usepackage{color,soul}
\usepackage{cite}
\usepackage{hyperref}
\usepackage[pscoord]{eso-pic}
\title{ \LARGE \bf AVGCN: Trajectory Prediction using Graph Convolutional Networks Guided by Human Attention}

\author{Congcong Liu$^{*}$, \  Yuying Chen$^{*}$,   \  Ming Liu,  \  Bertram E. Shi 
\thanks{* These two authors contributed equally}
\thanks{All the authors are with the Hong Kong University of Science and Technology. {\tt\small \{cliubh, ychenco, eelium, eebert\}@ust.hk}}  %
}

\begin{document}

\maketitle
\thispagestyle{empty}
\pagestyle{empty}
\begin{abstract}

Pedestrian trajectory prediction is a critical yet challenging task, especially for crowded scenes. 
We suggest that introducing an attention mechanism to infer the importance of different neighbors is critical for accurate trajectory prediction in scenes with varying crowd size.
In this work, we propose a novel method, AVGCN, for trajectory prediction utilizing graph convolutional networks (GCN) based on human attention (A denotes attention, V denotes visual field constraints). 
First, we train an attention network that estimates the importance of neighboring pedestrians, using gaze data collected as subjects perform a bird's eye view crowd navigation task. 
Then, we incorporate the learned attention weights modulated by constraints on the pedestrian's visual field into a trajectory prediction network that uses a GCN to aggregate information from neighbors efficiently. 
AVGCN also considers the stochastic nature of pedestrian trajectories by taking advantage of variational trajectory prediction.
Our approach achieves state-of-the-art performance on several trajectory prediction benchmarks, and the lowest average prediction error over all considered benchmarks.

\end{abstract}



\section{Introduction}

With the advancement of computer vision, trajectory forecasting has drawn much attention due to its essential applications in surveillance, crowd simulation, mobile robot navigation and autonomous driving. For applications in crowd analysis, it helps to understand the crowd behavior. 
For robots, understanding the future behavior of other agents better prepares it for valid and efficient planning and control.

Trajectory prediction is a challenging task. 
The trajectory can be influenced by multiple factors, including individual moving style, the underlying goal or intent, the environmental structure and topology, and the motion of other agents, etc. 
Pedestrian trajectory prediction is even more challenging, as humans have more freedom of motion than cars or robots. The pedestrian trajectories contain more frequent interactions and embody the stochastic properties of human motion. Furthermore, to tackle the higher density of human crowds, an efficient and effective approach is expected, considering the limited computation power and space of mobile devices. 

Work focusing on generating precise deterministic trajectory predictions have performed well \cite{alahi2016social}\cite{zhang2019sr}. However, they are not consistent with the stochastic nature of human motion. Recent work has highlighted the importance of stochasticity in trajectory prediction \cite{gupta2018social,sadeghian2019sophie,kosaraju2019social,lee2017desire,rhinehart2018r2p2}. Generative adversarial models (GANs) have been widely used to generate diverse predictions. 
However, adversarial training leads to unstable performance of GANs, which are highly sensitive to hyperparameters and changes in structure \cite{wang2019generative}. 
An alternative, the variational autoencoder (VAE), shows much more stable performance and has been applied to both vehicle \cite{lee2017desire} and trajectory prediction\cite{ivanovic2019trajectron,co2018self,felsen2018will}. 
In this paper, we exploited VAE-like architecture to capture stochasticity in trajectories.

Recent work has begun to consider the effect of other people on an individual trajectory. The fundamental problem is how to model the pairwise interactions into account and to adapt to varying crowd sizes. Some work has proposed to aggregate information of other agents through pooling method like max-pooling \cite{gupta2018social} and self-attention pooling \cite{sadeghian2019sophie}\cite{amirian2019social}. Other methods also utilized geometric information to model pedestrians in the local region with an occupancy map \cite{alahi2016social,bisagno2018group}. 
These methods either rely on some hand defined rules and tricks or only consider a subset of the pedestrians, which may cause information loss. As an alternative, some work proposes to represent connections among the pedestrians using graphs, e.g. spatial temporal graphs \cite{vemula2018social,ivanovic2019trajectron}. 
Graph convolutional network (GCN) \cite{chen2020robot,chen2020comogcn}, graph attention network (GAT) \cite{huang2019stgat,kosaraju2019social} and message passing neural network \cite{hu2020collaborative} have also been used to aggregate information about interactions. Inspired by \cite{chen2020robot}, we propose to use GCNs to aggregate interaction information. This requires many fewer parameters than GATs and places no limitations on the number of pedestrians. Another advantage of GCNs is that the interactions among the pedestrians can be conveniently modulated through the adjacency matrix. 
This raises the problem of how to model interactions in a large crowd of pedestrians. Our approach is based on the human use of attention to deal with the huge amount of visual information, as well as studies that have shown that visual field constraints influence human trajectories \cite{hasan2018seeing}. Pedestrians in front are more likely to be noticed while pedestrians in back are more likely to be ignored. 
In addition to visual field constraints, the importance of other pedestrians can also be influenced by other factors, such as the moving pattern. 
Our approach takes both the relative position as well as the motion of the pedestrians into account. 





In summary, this work addresses the aforementioned challenges in two ways. First, we use a graph structure to represent crowd state. Second, we use gaze data from human operators performing a bird-view navigation task to learn a network that assigns different weights to different pedestrians in the crowd according to their importance as measured by attention. 
We further constrain the weights using the visual field constraints of each pedestrian. 

There are several key contributions of our work:
\begin{itemize}

\item We exploit graph convolutional networks (GCN) to better model social interactions within human crowds. The use of GCN enables our approach to adapt to varying crowd sizes in a principled way. 

\item We exploit human attention to guide attention assignment. Because the bird's eye view we use to collect gaze data is not available to pedestrians actually in the crowd, to get closer to real world situations, we take visual field constraints into consideration.

\item We take advantage of variational inference to model the stochasticity of trajectories. 

\item With the above mechanisms, our AVGCN achieves state-of-the-art performance in several different trajectory prediction benchmarks, and the best average performance across all datasets considered.    

\end{itemize}

\section{Methods}
\subsection{Problem Formulation}
\label{sec:formulation}
The goal of our work is to generate probable future trajectories of all pedestrians in a scene.
The trajectory of a pedestrian $i$ is defined as $\boldsymbol{x_{i}^{(t_{s}:t_{e})}} = [(\text{x}_i^{t_{s}}, \text{y}_i^{t_{s}}), ..., (\text{x}_i^{t_{e}}, \text{y}_i^{t_{e}})]$, which denotes the position sequence of pedestrian $i$ from time stamp $t_s$ to $t_e$. $\boldsymbol{x_{\text{rel}_{i}}^t} = \boldsymbol{x_{i}^{t}} - \boldsymbol{x_{i}^{t-1}}$ denotes the change in position of pedestrian $i$ from time stamp $t-1$ to $t$. 
$\boldsymbol{p^{t_{\text{obs}}}_{\text{rel}_{i}}}[j] = \boldsymbol{x_{j}^{t_\text{obs}}} - \boldsymbol{x_{i}^{t_\text{obs}}}$ denotes the relative displacement from pedestrian $j$ to $i$ at time step $t_{obs}$. 
Consistent with \cite{sadeghian2019sophie,gupta2018social}, we express the input trajectory of all humans in a scene by $\boldsymbol{x_{\text{rel}_{1,...,N}}^{(1:t_{\text{obs}})}}$ for time step $t = 1,...t_{\text{obs}}$. The ground truth future trajectory is denoted by $\boldsymbol{x_{\text{rel}_{1,...,N}}^{(t_{\text{obs}}+1:t_{\text{\text{obs}}}+T)}}$ for time step $t = t_{\text{obs}} + 1,..., t_{\text{obs}} + T$. $N$ is the number of pedestrians. We also use the relative displacements from all pedestrians to the interested pedestrian, e.g. $\boldsymbol{p^{t_{\text{obs}}}_{\text{rel}_{i}}}{[1:N]}$, as input to better capture the social context of the pedestrian. 

The model generates predicted trajectories $\boldsymbol{\hat{x}_{\text{rel}_{1,...,N}}^{(t_{\text{obs}}+1:t_{\text{obs}}+T)}}$ that match the ground truth future trajectories $\boldsymbol{x_{\text{rel}_{1,...,N}}^{(t_{\text{obs}}+1:t_{\text{obs}}+T)}}$.

\subsection{Overall Model}
\begin{figure*}[htb]
\centering
\includegraphics[width=\textwidth]{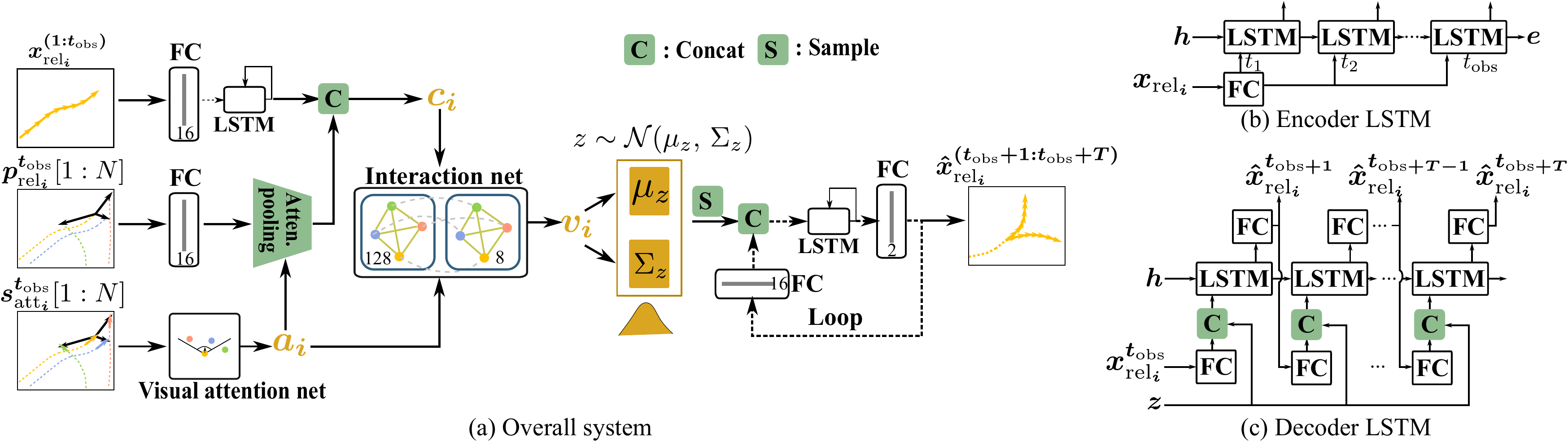}
\caption{The overall system for trajectory prediction. For clarity, we show only the trajectory prediction process for pedestrian $i$. The system contains a GCN-based variational encoder-decoder backbone for sequence-to-sequence trajectory prediction. For each pedestrian, an attention network assigns attention to neighboring pedestrians depending on their position relative to pedestrian $i$ and their velocities. Then, a visual field filter modulates the attention according to real-world visual field constraints. The resulting attention weights are applied to attention pooling and to modulating the adjacency matrix of GCN. The sequence-to-sequence prediction is realized by two LSTMs. In (b) and (c), we unroll the LSTM to show the mapping from input to output clearly. }
\label{fig:system_arc}
\vspace{-1.5em}
\end{figure*}
Figure \ref{fig:system_arc} shows the overall framework of our approach for pedestrian trajectory prediction. 
For ease of exposition, we show the prediction process for pedestrian $i$ as an example. The process for other pedestrians is similar. 

As discussed in section \ref{sec:formulation}, we capture the individual's motion through the sequence of relative positions $\boldsymbol{x_{\text{rel}_{i}}^{(1:t_{\text{obs}})}}$ and represent the social context by the relative displacements $\boldsymbol{p^{t_{\text{obs}}}_{\text{rel}_{i}}}{[1:N]}$ of all pedestrians to pedestrian $i$ at time $t_{\text{obs}}$. 

To represent an individual's trajectory as a fixed length vector, we first apply a single layer MLP (FC) to the each of relative positions. We then use an LSTM to process the sequence over time, i.e 
\begin{equation}
\label{eqn:encodelstm}
\boldsymbol{e_i}=LSTM_{\text{en}}(MLP_{\text{mot}}(\boldsymbol{x_{\text{rel}_i}^{1:t_{\text{obs}}}};W_{\text{mot}}),\boldsymbol{h_{\text{enc}_i}};W_{\text{en}})
\end{equation}
where $W_{\text{mot}}$ denotes weight of single layer MLP, and $W_{\text{en}}$ denotes weight of the encoding LSTM. The final feature vector $\boldsymbol{e_i}$ is the hidden state vector of the LSTM at time $t_\text{obs}$.

Similarly, the relative displacements of all humans are fed into an FC layer first. Then an attention pooling module computes a weighted sum of the embeddings across the different people to get the combined social context feature $\boldsymbol{p_i}$ for pedestrian $i$.
\begin{equation}
\label{eqn:encodelstm}
\boldsymbol{p_i}= MLP_{\text{cont}}(\boldsymbol{p^{t_{\text{obs}}}_{\text{rel}_{i}}}{[1:N]};W_{\text{cont}}) \boldsymbol{\cdot} \boldsymbol{a_i}
\end{equation}
where $W_{\text{cont}}$ denotes weight of single layer MLP, and $\boldsymbol{a_i}$ is a vector that denotes the attention weights assigned to each pedestrian. 

The motion pattern feature $\boldsymbol{e_i}$ and social context feature $\boldsymbol{p_i}$ are concatenated as the input feature $\boldsymbol{c_i}$ for pedestrian $i$. We deploy a two-layer GCN to integrate information across all pedestrians in the crowd, simultaneously to obtain $N$ social interaction features $\boldsymbol{v_1}, ..., \boldsymbol{v_N}$.
\begin{align}
\label{eqn:GCN}
[\boldsymbol{v_1}; \boldsymbol{...}; \boldsymbol{v_N}] &= GCN_{\text{si}}([\boldsymbol{c_1}; \boldsymbol{...}; \boldsymbol{c_N}]; \boldsymbol{A_{\text{si}}}, W_{\text{si}}) \\
&\boldsymbol{A_{\text{si}}} = [\boldsymbol{a_1}; \boldsymbol{...}; \boldsymbol{a_N}] 
\end{align}

Each pedestrian corresponds to one node of the graph. The $i^\text{th}$ row of the adjacency matrix $\boldsymbol{A_{\text{si}}}$ contains the attention vector containing the attention paid by pedestrian $i$ to each pedestrian in the crowd, where the subscript \text{si} stands for social interaction. $W_{\text{si}}$ denotes the weight matrices of the two-layer GCN. The use of a two layer GCN enables us to capture more complex interactions than simple pairwise interactions.


Social interaction features computed by the GCNs were used to estimate the means and variances of a distribution over trajectory features using a pair of MLPs. 
\begin{equation}
\label{eqn:vae}
\boldsymbol{\mu_z}, \boldsymbol{\Sigma_z}=MLP_{\text{mean}}(\boldsymbol{v_i}; W_{\text{mean}}), MLP_{\text{var}}(\boldsymbol{v_i}; W_{\text{var}})
\end{equation}

where $W_{\text{mean}}$ and $W_{\text{var}}$ denote the weight matrices. 
We sample a trajectory feature $\boldsymbol{z}$ from this distribution $\boldsymbol{z\sim\mathbf{N}(\mu_z,\Sigma_z)}$.
This trajectory feature is used to generate a predicted trajectory using a decoder LSTM connected in a feedback configuration (Fig. \ref{fig:system_arc}(c)). At each prediction time step $t \in [t_{\text{obs}+1}, ..., t_{\text{obs}+T}]$, the trajectory feature $\boldsymbol{z}$ is concatenated with an embedding of the predicted location at time $t-1$, $MLP_{\text{enc}} (\boldsymbol{\hat{x}_{\text{rel}_i}^{t-1}})$ before input to the LSTM. The hidden fearure of the decoder LSTM is input to a decoding MLP, $MLP_\text{dec}$ to generate a trajectory prediction:

\begin{equation}
\label{eqn:decodelstm}
\begin{split}
\boldsymbol{\hat{x}_{\text{rel}_i}^{t}}=&MLP_{\text{dec}}(LSTM_{\text{de}}(concat(\boldsymbol{z},\\& MLP_{\text{enc}}(\boldsymbol{\hat{x}_{\text{rel}_i}^{t-1}}, W_\text{enc})),  \boldsymbol{h_{\text{de}_i}};W_{\text{de}});W_{\text{dec}})
\end{split}
\end{equation}
where $W_{\text{enc}}$, $W_{\text{de}}$ and $W_{\text{dec}}$ indicate weight parameters.

We minimize the following loss for trajectory prediction:  
\begin{equation}
   L_{\text{pred}} = ||\boldsymbol{x_{\text{rel}_i}} - \boldsymbol{\hat{x}_{\text{rel}_i}}||+ \alpha KL(\mathbf{z}, \mathbf{p}).
\end{equation}
where $\mathbf{p}\sim\mathbf{N(0,1)}$ is a prior distribution and $\alpha$ is a weighting factor.
The KL loss forces the distribution of the mean and covariance matrix in (\ref{eqn:vae}) to be close to a normal distribution.   

The attention weights are used for attention pooling and for constructing the adjacency matrix. We used a visual attention module, introduced in detail in section \ref{sec:atten}, to obtain these attention weights.

\subsection{Attention in Trajectory Prediction}
\label{sec:atten}
\subsubsection{Attention Estimation}
\begin{figure}[!htb]
\centering
\includegraphics[width=0.8\columnwidth]{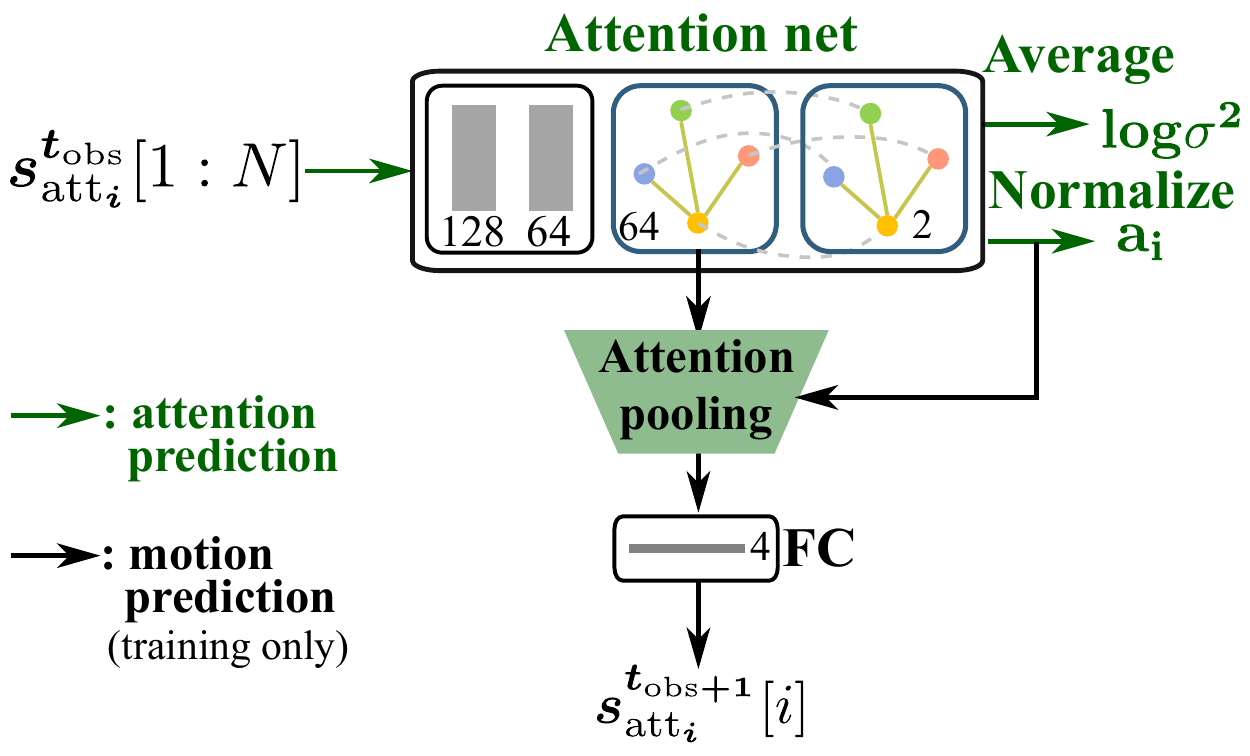}
\caption{The structure of the network estimating attention for pedestrian $i$. The network for other pedestrians is similar. The attention weights are learned to optimize performance on a motion prediction task.}
\label{fig:attention_arc}
\vspace{-0.5em}
\end{figure}
In previous work, human gaze, has been used as a proxy of attention. Chen \textit{et al.} \cite{chen2020robot} learned attention directly by minimizing an attention estimation error defined based on gaze data. They did not take into account any task specific demands. In contrast, we combine motion prediction error with attention estimation error, shown in Fig. \ref{fig:attention_arc}.

We adopt a two-layer GCN to estimate the attention weights pedestrian $i$ assigns to each pedestrian including him/herself. The predicted attention weights are used to pool the outputs of the first GCN layer. The pooled output feature is then used to predict the motion of pedestrian $i$, using a fully connected single layer neural network. 

The input to the attention network are the positions of the pedestrians in the crowd relative to pedestrian $i$ and their velocities at time $t_{\text{obs}}$.
\begin{equation}
\label{eqn:motion}
  \boldsymbol{s^{t_{\text{obs}}}_{\text{att}_i}}[j] = [\text{x}_j^{t_{\text{obs}}} - \text{x}_i^{t_{\text{obs}}} , \text{y}_j^{t_{\text{obs}}} - \text{y}_i^{t_{\text{obs}}}, \text{v}_{x_j}^{t_{\text{obs}}}, \text{v}_{y_j}^{t_{\text{obs}}}].
\end{equation}

For $j \in \{1,\ldots,N \}$, we first embed these features into a higher dimensional space using a two layer MLP:
\begin{equation}
  \boldsymbol{b_i}{[1:N]} = MLP_{\text{att}}(\boldsymbol{s^{t_\text{obs}}_{\text{att}_i}}{[1:N]}; W_{\text{att}}).
\end{equation}
where $W_{\text{att}}$ denotes the weights.


The embeddings are fed into a two-layer GCN, which produces two outputs per node.
\begin{align}
\label{eqn:att_gcn}
[q_{i1}, \mathrm{a_{i1}}; ...; q_{iN}, \mathrm{a_{iN}}]= GCN_{\text{att}}([\boldsymbol{b_i^1}; \boldsymbol{...}; \boldsymbol{b_i^N}]; \boldsymbol{A_{\text{att}}},  W_{\text{gcn}}).
\end{align}
where $W_{\text{gcn}}$ denotes the weights.

We set the adjacency matrix $\boldsymbol{A_{\text{att}}}$ according to a star topology, where the pedestrian $i$ is the central connection point \cite{kipf2016semi}. 
Rows are normalized to sum to one.
The output feature dimension of each node in the first graph convolutional layer is set to 64. The output dimension of each node in the second layer is two. 

The weights in the attention network minimize a loss function that computes a motion prediction error and a divergence from the ground truth gaze.
\begin{equation}
    \label{eqn:loss}
   L_{\text{att}} = ||\boldsymbol{s^{t_{\text{obs}+1}}_{\text{att}_{i}}}[i] - \boldsymbol{\hat{s}^{t_{\text{obs}+1}}_{\text{att}_{i}}}[i]||_2^2+ \beta KL(\mathbf{a_i}, \boldsymbol{a_{\text{gt}}}(\sigma)).
\end{equation}
where $\beta$ is a weighting factor in the combination. We describe this two components of the loss function in more detail below.

$\mathbf{a_i}$ in the second term is the estimated attention weights of pedestrian $i$ to other nodes.
\begin{equation}
    \mathbf{a_i} =  \text{Normalize}([\mathrm{a_{i1}}; ...; \mathrm{a_{iN}}])
\end{equation}
The ground truth attention weights $\boldsymbol{a_{\text{gt}}}(\sigma)$ are calculated from the gaze points collected in the 0.2 seconds time window at $t_{\text{obs}}$. We first created a Gaussian mixture model by placing Gaussians with variance $\sigma^2$ at each gaze point. The attention weights for pedestrians are obtained by directly obtaining the Gaussian mixture density at their positions and then normalized to sum to one. For simplicity, we use $x_g$ to denote the positions of gaze data.

\begin{equation}
\label{eqn:gt_atten}
   a_{\text{gt}_j}(\sigma) =\frac{e^{(x_{p_j}-x_g)^2/2\sigma^2}}{\sum_j  e^{(x_{p_j}-x_g)^2/2\sigma^2} }
\end{equation}
where $j \in \{ 1,\ldots,N \}$, $x_{p_j}$ is the position of pedestrian $j$. 

The ground truth attention weights depend on the variance $\sigma^2$. We estimated the log variance of Gaussian $log\sigma^2$ by averaging the first outputs of each node in (\ref{eqn:att_gcn}). 
It is different with the setting in \cite{chen2020robot}, where they fixed the $\sigma^2$ to obtain ground truth attention. However, this parameter should be environment-dependent. For example, in dense environments, we focus more on the local area to carefully avoid collisions, which indicates smaller $\sigma^2$ than that of a sparse environment. We embeded this parameter into our learning framework to avoid choosing it in an ad-hoc manner.

\begin{equation}
    log\sigma^2 =  \text{Average}([q_{i1}; ...; q_{iN}])
\end{equation}

   

To compute the first term in (\ref{eqn:loss}), we estimated the next frame motion $\boldsymbol{{s}^{t_{\text{obs}+1}}_{\text{att}_{i}}}[i]$ of pedestrian $i$, where $\boldsymbol{{s}^{t_{\text{obs}+1}}_{\text{att}_{i}}}[i] = [\text{x}_i^{t_{\text{obs}}+1} - \text{x}_i^{t_{\text{obs}}} , \text{y}_i^{t_{\text{obs}}+1} - \text{y}_i^{t_{\text{obs}}}, \text{v}_{x_i}^{t_{\text{obs}}+1}, \text{v}_{y_i}^{t_{\text{obs}}+1}]$.
We generated $\boldsymbol{\hat{s}^{t_{\text{obs}+1}}_{\text{att}_{i}}}[i]$ from a single FC layer that takes the embedding after the first graph convolutional layer and multiplied by the estimated attention weights as input. 
\begin{equation}
   \boldsymbol{\hat{s}^{t_{\text{obs}+1}}_{\text{att}_{i}}}[i] = FC(\boldsymbol{u_i}{[1:N]} \boldsymbol{\cdot} \mathbf{a_i}; W).
\end{equation}
where $\boldsymbol{u_i}{[1:N]} \in \mathbb{R}^{N \times 64}$ is the embedding after the first graph convolutional layer.





\subsubsection{Pedestrian Visual Field Constraints}
The gaze data used to estimate the ground truth attention was collected from human operators performing a robot navigation task through a crowd based on a top-down view of the environment. Thus, the operators had access to a different perspective than real-world pedestrians, whose visual field is limited to the area in front of them. To compensate for this mismatch, 
we added a visual field constraint to the estimated attention weights, which assumed that each pedestrian only pays attention to pedestrians located within a certain visual angle of the front facing direction of pedestrian $i$ as determined by the velocity as shown in the bottom of Fig. \ref{fig:system_arc}.

We swept the visual field angle and found that 120-degrees of visual field yields best performance on the validation set, which is consistent with the horizontal visual angle for binocular vision \cite{henson2000visual}.

We refer to the concatenation of attention network with the visual angle filter as the visual attention net. The attention generated by the visual attention net is:
\begin{equation}
    \boldsymbol{a_i} = \text{VisualFilter}(\mathbf{a_i})
\end{equation}
which is applied for combing the social context feature in (2) and obtaining social interaction feature in (3)-(4).

\subsection{Implementation details}
To learn the attention network, we used the Adam optimizer with the learning rate 1e-3. The modes were trained for 100 epochs. The activation function is the ReLU. The values of $\beta$ was 0.5 which yields lowest motion prediction error over validation dataset. 
The dimension of two-layer MLP ($MLP_{\text{att}}$) is 128 follow by 64. The hidden number of two-layer GCN ($GCN_{\text{att}}$) has the dimension of 64 and 8 separately. And the single FC layer ($FC$) has a dimension of 4 for motion prediction. For the data collection, we ask subjects to steer a “virtual pedestrian”, through a crowd based on a top-down view of the scene. The crowd is constructed with the real human pedestrian trajectories that used for training the pedestrian trajectory prediction network. As the human performs this task, we monitor his or her gaze. The gaze forms the ground-truth that we use to train the attention model and is only used for the training. 

For pedestrian trajectory prediction network, we trained the network with Adam optimizer. The models were trained for 200 epochs. The mini-batch size is 64 and the learning rate is 1e-4. The $\alpha$ is 0.001 which yields lowest motion prediction error over validation dataset.
The encoder encodes the relative trajectories by a single layer of MLP ($MLP_{\text{enc}}$) with a dimension of 16 followed by an LSTM ($LSTM_{\text{en}}$)with a hidden dimension of 32. The embedding output from LSTM was then concatenated with the features extracted from the relative position from other humans by a single MLP with dimension of 16.
The concatenated features are then fed into two-layer GCNs for feature integration.
Then hidden number for two graph convolutional layer has the dimension of 128 and 8 separately.
Then MLPs ($MLP_{\text{mean}}$, $MLP_{\text{var}}$) were used to take state of humans to create a distribution with mean and variance. Then we sample $\boldsymbol{z}$ from this distribution with a dimension of 8, and fed it into an LSTM ($LSTM_{\text{de}}$) with dimension of 32 and followed by an MLP ($MLP_{\text{dec}}$)with dimension of 2 for decoding. We used the same training, validation, and testing dataset as S-GAN\cite{gupta2018social}.

\section{Experiments}

\subsection{Evaluation methodology}
Following the setting in \cite{gupta2018social}, we adopt the leave-one-out approach, i.e., train with four datasets and test in the remaining dataset.
We take trajectories with 8 time steps as observation and evaluate trajectory predictions over the next 12 time steps.
\subsubsection{Metrics}
Consistent with previous works \cite{gupta2018social,sadeghian2019sophie,kosaraju2019social}, we adopt two standard metrics in meter: Average Displacement Error (ADE) and Final Displacement Error (FDE).
\textit{ADE} denotes average L2 distance between ground truth and predictions of all time steps.
\textit{FDE} denotes average L2 distance between ground truth and prediction at the final time step.

\subsubsection{Baselines}
We compare our work with following several recent works based on generative models, including \textit{Social GAN (S-GAN)} \cite{gupta2018social}, \textit{Sophie} \cite{sadeghian2019sophie}, \textit{Trajectron} \cite{ivanovic2019trajectron}, \textit{Social-BiGAT (S-BiGAT)} \cite{kosaraju2019social}, \textit{CoMoGCN} \cite{chen2020comogcn}, \textit{STGAT} \cite{huang2019stgat}, and \textit{NMMP} \cite{hu2020collaborative}. 
%
%









\subsection{Attention network evaluation}

Consistent with the evaluation method of trajectory prediction, we adopt the leave-one-out approach for attention network evaluation, i.e. train with four datasets and test in the remaining dataset.

\begin{figure*}[!htb]
  \centering
  \includegraphics[width=1.4\columnwidth]{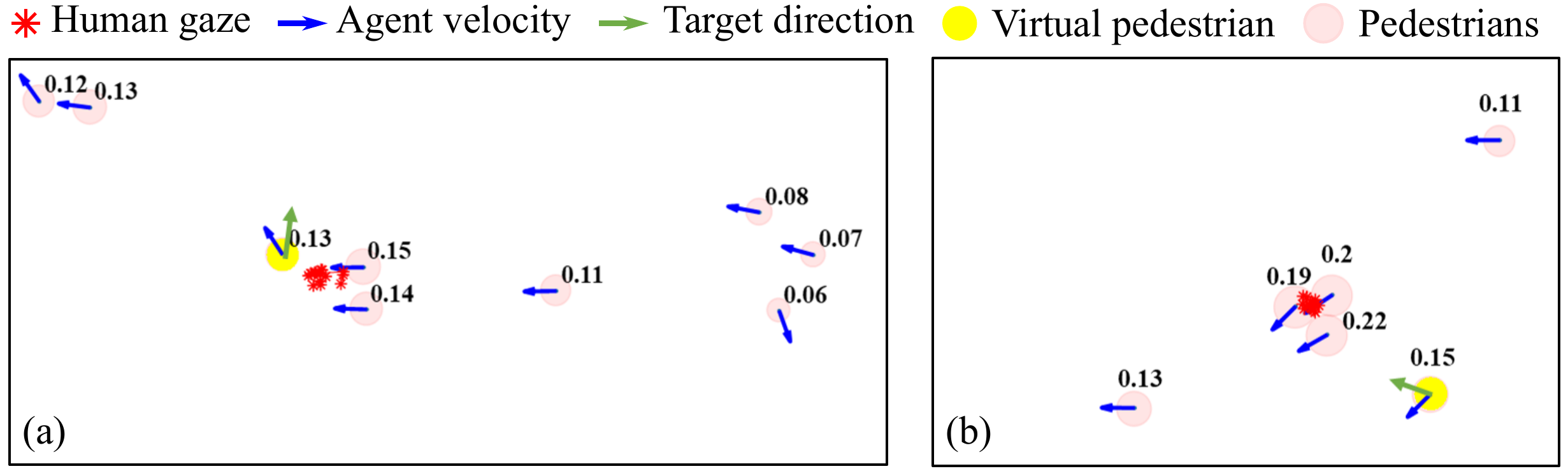}
  \caption{Two examples of estimated attention weights from the attention networks. The weights are shown by radius of the pink circle around each pedestrian and the value of weights are also marked around. Red stars show the gaze data used to compute ground truth attention weights. The virtual pedestrian controlled by subject is shown in yellow. Blue arrows show the instantaneous velocity of each pedestrian. Green arrow in the virtual pedestrian show the direction towards goal. Human subjects navigate the virtual pedestrian to the goal across crowds.
  } \label{fig:attention} 
  \vspace{-1em}
\end{figure*}

Fig. \ref{fig:attention} show two examples of the estimated attention weights with gaze data of the human subject in testing datasets.
As shown in Fig. \ref{fig:attention} (a), there are two pedestrians walking in front of it and also two pedestrians behind very close to the virtual pedestrian. To avoid collision, the human subject steer the virtual pedestrian to left side and the human subject's gaze is located on the two close pedestrians behind. The leaned attention weights are largest for the two close pedestrians.
For the scene in Fig. \ref{fig:attention} (b), the virtual pedestrian is going left while three pedestrians block the way. The human subject navigates the virtual pedestrian backward to avoid potential collisions, and the gaze is focused on the pedestrians. The learned attention weights are also largest for the three pedestrians.
Those qualitative results show that the attention network has learned to correctly infer which pedestrians inside crowds are more crucial for trajectory prediction. 


To evaluate the benefit of learned attention weights quantitatively, we compare the motion prediction performance between the model which considers the attention loss and the one without attention loss (i.e., set $\beta$ equals to 0). 

\begin{table}[!htb]
\centering
\begin{tabular}{|l|cc|}
\hline
Dataset & \multicolumn{1}{c|}{\begin{tabular}[c]{@{}c@{}}with \\ attention loss\end{tabular}} & \begin{tabular}[c]{@{}c@{}}without \\ attention loss\end{tabular} \\ \hline
ETH     & \textbf{0.130}                                                                    & 0.141                                                         \\
HOTEL   & \textbf{0.075}                                                                    & 0.081                                                         \\
UNIV    & \textbf{0.075}                                                                    & 0.083                                                         \\
ZARA1   & \textbf{0.088}                                                                    & 0.096                                                         \\
ZARA2   & 0.059                                                                             & \textbf{0.058}                                                \\ \hline
AVG     & \textbf{0.085}                                                                    & 0.092                                                         \\ \hline
\end{tabular}
\caption{Quantitative results for motion prediction. We use mean absolute error for evaluation over five different datasets. Without attention loss corresponds to the model use l2 motion prediction loss only, i.e., set $\beta=0$. We can see that learning attention as an additional task improve the motion prediction accuracy (Lower value suggest better performance).}
\label{tab:motion_pred}
\vspace{-1em}
\end{table}

As shown in Table \ref{tab:motion_pred}, the mean absolute error of motion prediction has been calculated over five different testing datasets. Comparing to the one not considering attention loss, we can see learning attention leads to 7.6\% improvement of prediction accuracy on average. This suggests that modulating the embeddings with learned attention weights helps the network to more accurately predict the next frame motion.
In the next section, we will show the benefits of the learned attention weights for the multiple pedestrian prediction task.

\subsection{Trajectory prediction with learned attention}

\subsubsection{Quantitative Evaluation}

\begin{table*}[!htb]
\centering
\begin{tabular}{|l|ccccccc|c|l|l}
\hline
        & \multicolumn{7}{c|}{Baselines}                                                                                                                                                                      & \multicolumn{3}{c|}{Ours}        \\ \cline{2-11} 
Dataset & \multicolumn{1}{c|}{S-GAN} & \multicolumn{1}{c|}{Sophie} & \multicolumn{1}{c|}{Trajectron} & \multicolumn{1}{c|}{S-BiGAT} & \multicolumn{1}{c|}{CoMoGCN} & \multicolumn{1}{c|}{STGAT} & NMMP        & \multicolumn{3}{c|}{AVGCN}       \\ \hline
ETH     & 0.81/1.52                  & 0.70/1.43                   & 0.59/1.17                       & 0.69/1.29                    & 0.70/1.28                    & 0.70/1.21                  & 0.67/1.22   & \multicolumn{3}{c|}{\textbf{0.62/1.06}}   \\
HOTEL   & 0.72/1.61                  & 0.76/1.67                   & 0.42/0.80                       & 0.49/1.01                    & 0.37/0.75                    & 0.32/0.63                  & 0.33/0.64   & \multicolumn{3}{c|}{\textbf{0.31/0.58}}   \\
UNIV    & 0.60/1.26                  & 0.54/1.24                   & 0.59/1.21                       & 0.55/1.32                    & 0.53/1.16                    & 0.56/1.20                  & \textbf{0.52/1.12}   & \multicolumn{3}{c|}{0.55/1.20}   \\
ZARA1   & 0.34/0.69                  & 0.30/0.63                   & 0.55/1.09                       & \textbf{0.30/0.62}                    & 0.34/0.71                    & 0.33/0.64                  & 0.32/0.66   & \multicolumn{3}{c|}{0.33/0.70}   \\
ZARA2   & 0.42/0.84                  & 0.38/0.78                   & 0.52/1.04                       & 0.36/0.75                    & 0.31/0.67                    & 0.30/0.61                  & 0.29/0.62   & \multicolumn{3}{c|}{\textbf{0.27/0.58}}   \\ \hline
AVG     & 0.58/1.18                  & 0.54/1.15                   & 0.53/1.06                       & 0.48/1.00                    & 0.45/0.91                    & 0.44/0.86                  & 0.426/0.852 & \multicolumn{3}{c|}{\textbf{0.416/0.824}} \\ \hline
\end{tabular}
\caption{Quantitative results of trajectory prediction for comparison with state-of-the-art. We adopted two metrics Average Displacement Error (ADE) and Final Displacement Error (FDE) in meters for evaluation over five different datasets. On average, our full model (AVGCN) achieves state-of-the-art results outperforming all baseline methods.} 
\label{tab:pred_error_baseline}
\vspace{-1em}
\end{table*}

\textit{Comparison to state-of-the-art methods}:
As shown in Table \ref{tab:pred_error_baseline}, we compare our models with various baselines. The average displacement error (ADE) and final displacement error (FDE) were reported across five datasets.
Following settings in every baseline, we run 20 samples for evaluation.

\begin{table}[!htb]
\centering
\resizebox{\columnwidth}{!}{
\begin{tabular}{|l|cccc|}
\hline
        & \multicolumn{4}{c|}{Ablation study}                                                            \\ \cline{2-5} 
Dataset & \multicolumn{1}{c|}{GCN} & \multicolumn{1}{c|}{AGCN} & \multicolumn{1}{c|}{VGCN} & AVGCN       \\ \hline
ETH     & 0.72/1.41                & 0.77/1.35                 & 0.69/1.26                 & \textbf{0.62/1.06}   \\
HOTEL   & 0.37/0.71                & \textbf{0.30}/0.58                 & 0.33/\textbf{0.56}                 & 0.31/0.58   \\
UNIV    & 0.64/1.34                & \textbf{0.55/1.18}                 & 0.63/1.33                 & \textbf{0.55}/1.20   \\
ZARA1   & 0.34/0.71                & \textbf{0.33/0.70}                 & 0.35/0.70                 & \textbf{0.33/0.70}   \\
ZARA2   & 0.30/0.62                & \textbf{0.26/0.56}                 & 0.27/0.57                 & 0.27/0.58   \\ \hline
AVG     & 0.474/0.958              & 0.44/0.872                & 0.454/0.884               & \textbf{0.416/0.824} \\ \hline
\end{tabular}
}
\caption{Quantitative results for ablation study. We adopted two metrics Average Displacement Error (ADE) and Final Displacement Error (FDE) in meters for evaluation over five different datasets. Lower value denotes better performance.}
\label{tab:pred_error_ablation}
\vspace{-1.5em}
\end{table}

\begin{figure*}[!htbp]
\centering
\includegraphics[width=0.8\textwidth]{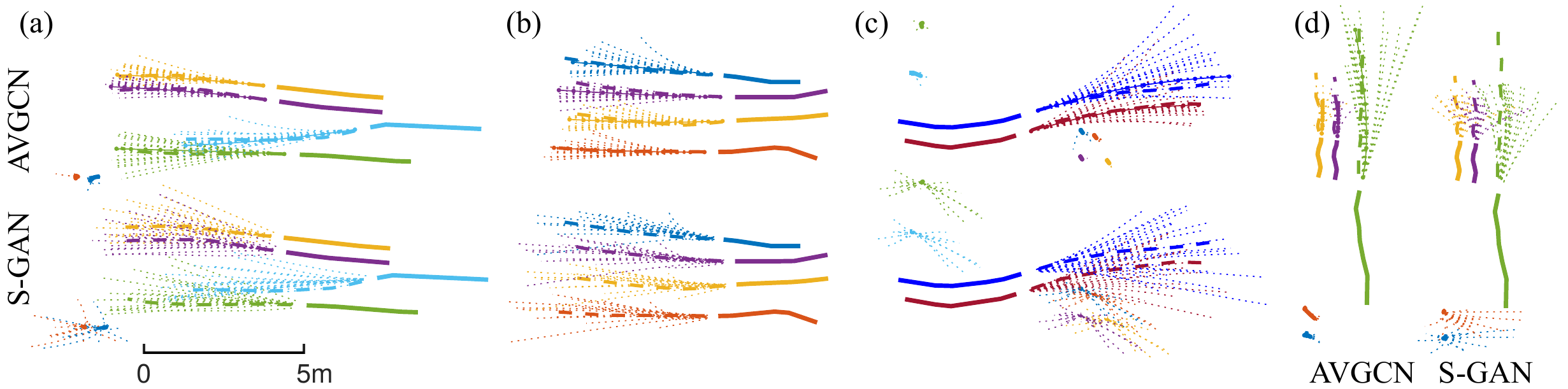}
\caption{Examples for generated trajectories with our model AVGCN and S-GAN. The dotted lines show the twenty stochastically generated trajectory samples. The solid lines show the historical trajectory observed. The dashed lines show the ground-truth future trajectory. We also represent the ``average" prediction of variational encoder-decoder by applying the mean value ($\mu_z$) of the distribution. They are shown in dot-solid lines. }
\label{fig:quali_analysis}
\vspace{-1em}
\end{figure*}
It is clear to see that our final model with GCN and attention learned from gaze data beat all baselines with least average ADE/FDE and obtain more consisting results in both ADE and FDE. 
Compared to S-GAN, we achieve 28.3\% improvement in ADE and 30.2\% improvement in FDE. 
Compared to Sophie who use additional scene context information, we achieve 23.0\% improvement in ADE and 28.3\% improvement in FDE. 
Compared to Trajectron who uses VAE as backbone network, we achieve 21.5\% improvement in ADE and 22.3\% improvement in FDE. 
Compared to CoMoGCN, we achieve 7.6\% improvement in ADE and 9.5\% improvement in FDE. 
Compared to STGAT, we achieve 5.5\% improvement in ADE and 4.2\% improvement in FDE. 
The result reported in the table is obtained from running their code in the default setting. 
Compared to NMMP, we achieve 2.3\% improvement in ADE and 3.3\% improvement in FDE. 
The result reported in the table is obtained from running their provided model. 
Compare to S-BiGAT who also considers graph structure for interaction modeling, we achieve 13.3\% improvement in ADE and 17.6\% improvement in FDE. 

\textit{Ablation study}: As shown in Table \ref{tab:pred_error_ablation}, We conduct several ablation studies to validate the benefits of the use of GCN and attention mechanisms.
\textit{Vanilla GCN (GCN)} uses vanilla GCN for crowd information aggregation without considering any attention mechanism.
\textit{AGCN} incorporated learned attention weights from human gaze by modulating the adjacency matrix of GCN.
\textit{VGCN} use visual field filter to modulate the adjacency matrix.
\textit{AVGCN} first use learned attention weights to modulated the adjacency matrix, then it add visual field constraint which only considers humans in front inside a visual angle.


To show the benefit of the incorporation of attention learned from human gaze, we compare two pairs of models (AGCN vs. GCN) and (AVGCN vs. VGCN). 
Incorporating attention from human gaze improves ADE by 8.4\% (AVGCN vs. VGCN) and 7.2\% (AGCN vs. GCN), and improves FDE by 6.8\% (AVGCN vs. VGCN) and 9.0\% (AGCN vs. GCN).

Furthermore, to validate the benefit of adding visual field constraints, we compare two pairs of models (AGCN vs. AVGCN) and (GCN vs. VGCN).
Adding visual field constraints improves ADE by 4.2\% (GCN vs. VGCN) and 5.5\% (AGCN vs. AVGCN), and improves FDE by 7.7\% (GCN vs. VGCN) and 5.5\% (AGCN vs. AVGCN).

The above ablation studies clearly demonstrate the benefits of the use of visual field constraints and the introduction of attention mechanisms.

\subsubsection{Qualitative Evaluation}

To better understand the benefits of our model, we visualize several examples of predicted trajectories in testing datasets as shown in Fig.\ref{fig:quali_analysis}.

Comparing to S-GAN, we can see that the trajectories generated by our model are more accurate and realistic. To be more specified, trajectories generated by S-GAN has larger variance when it is supposed to be constrained by the social interactions (the yellow and purple trajectories in Fig.\ref{fig:quali_analysis}(a) and (d)), while our model predict more reasonable trajectories(Fig.\ref{fig:quali_analysis}(c)). The distribution of the trajectory samples deviates from the ground truth more frequently (the green and blue trajectories in Fig.\ref{fig:quali_analysis}(a)). Also it can be observed that our method has smaller accumulating displacement error while the trajectory samples generated by S-GAN only cover partial of the ground-truth trajectory (Fig.\ref{fig:quali_analysis}(b) and (d)). Larger variance results in larger coverage of the predicted trajectories. However, it is not expected by downstream tasks like robot navigation. Better modeling of interactions can reduce the variances and better predict trajectories. This suggests that our model with better interactions modeling can generate trajectories more efficiently and accurately.

\section{Conclusion}
In this paper, we propose a novel VAE-like network AVGCN that utilizes attention mechanism for multiple pedestrian trajectory prediction. The proposed model outperforms the state-of-the-art methods over different datasets.
We introduced graph convolutional networks for efficient social interaction aggregation.
First we learned an attention network from human gaze in a motion prediction task. Then we incorporated the learned attention weights to modulate the adjacency matrix in GCN for pedestrian trajectory prediction. Furthermore, for better incorporate attention weights in a more realistic way, we considered extra visual field constraints after the attention modulation considering the limited visual field of human in real-world.
We show that the use of graph convolutional networks and attention mechanism significantly improve the performance of model for pedestrian trajectory prediction.

\bibliographystyle{IEEEtran}
\bibliography{example}
\end{document}